\documentclass[conference]{IEEEtran}
\IEEEoverridecommandlockouts
\usepackage{cite}
\usepackage{amsmath,amssymb,amsfonts}
\usepackage{algorithmic}
\usepackage{graphicx}
\usepackage{textcomp}
\usepackage{xcolor}
\usepackage{tabularx}
\usepackage{tabularray}
\usepackage{multirow}
\usepackage{multicol}
\def\BibTeX{{\rm B\kern-.05em{\sc i\kern-.025em b}\kern-.08em
    T\kern-.1667em\lower.7ex\hbox{E}\kern-.125emX}}
\begin{document}

\title{ViM-Disparity: Bridging the Gap of Speed, Accuracy and Memory for Disparity Map Generation}

\author{\IEEEauthorblockN{Maheswar~Bora\textsuperscript{*}\thanks{\textsuperscript{*}The first and second authors have equal contributions}, Tushar~Anand\textsuperscript{*}, 
Saurabh~Atreya,
Aritra~Mukherjee, 
Abhijit~Das}
\IEEEauthorblockA{Machine Intelligence Group, Department of CS\&IS, Birla Institute of Technology and Sciences, Pilani, India.
\\
Email: abhijit.das@hyderabad.bits-pilani.ac.in}}

\maketitle

\begin{abstract}
In this work we propose a Visual Mamba (ViM) based architecture, to dissolve the existing trade-off for real-time and accurate model with low computation overhead for disparity map generation (DMG). Moreover, we proposed a performance measure that can jointly evaluate the inference speed, computation overhead and the accurateness of a DMG model. The code implementation and corresponding models are available at: https://github.com/MBora/ViM-Disparity.

\end{abstract}
\section{Introduction}
Fast disparity map generation from stereo imagery, has always been one of the most challenging aspects of robotic vision \cite{hamzah2016literature, scharstein2002taxonomy}. The traditional trend of DMG had been in the form of patch matching from the left to the right image along the Epipolar plane. However, it has been observed that such assumptions do not hold for robotic applications, as, in such a scenario, the camera view keeps rotating and experiences vibrations~\cite{liang2018learning}. Thus in recent times, stereo~\cite{hamid2022stereo, mayer2016large, chang2018pyramid, khamis2018stereonet, song2019edgestereo, wang2019anytime, shen2022pcw, xu2021digging, huang2022h, teed2020raft, xu2023iterative, xu2023unifying} or monocular~\cite{ming2021deep, masoumian2022monocular, vishwanath2013seeing, mukherjee2023recent, rachalwar2023depth, tosi2019learning}, and even multi-sensor input-based \cite{6623451} deep learning models are explored for DMG. For models based on monocular depth estimations the precision is not as high as stereo-based DMG models \cite{sleaman2023monocular}. Whereas, in stereo-based models precision is high but the real-time inference and computational overhead are high which is not suitable for real-life applications. In contrast, there are works in the literature that focus in the direction of real-time inference \cite{dosovitskiy2020image, xu2022attention, chen2022improvement, liang2023, pan2024dtpnet} and reduce computation overhead for DMG \cite{yang2019hierarchical, khot2019learning, dai2019mvs2, pilzer2019progressive, tankovich2021hitnet, lipson2021raft}, but they lack in precision. To conclude, there is a gap in the literature for real-time and accurate models with low computation overhead for DMG. This motivates us to propose a model that can dilute the above mentioned tradeoff. 

Specifically, we proposed a VisionMamba~\cite{zhu2024vision} based model for faster DMG.  Moreover, the performance metrics that are widely used in literature for DMG are End Point Error (EPE) and 3-pixel error accuracy percentage for disparity (D1) are independent of speed and memory consumed by models. Hence, we proposed a performance measure SOMER that computes the ratio of the error with time and memory factor to find a unified measure for DMG model. 

The specific contribution of our works is as follows:
\begin{itemize}
    \item We proposed a new architecture based on ViM that performs fast DMG restoring accuracy with low memory requirement.
    \item We proposed a new metric of overall efficacy measurement of DMG methodologies as Speed Over Memory and Error Ratio (SOMER).
     \item A robust benchmarking of the recent DMG techniques available in the literature.

\end{itemize}

\section{Proposed Methodology}
After a critical study of emerging works in real-time disparity map computation, we observed that an attention-based model is quite suitable for such application~\cite{xu2023unifying}. Joint attention between extracted features from both left and right images, rather than the raw image itself has been explored in such models. Though attention models are quite heavy and thus take a considerable time to compute the disparity map for a pair of stereo frames, their accuracy is very robust. The recent introduction of Mamba~\cite{gu2023mamba}, a direct state space model (SSM) an alternative for transformers that can help in reducing the computation time for DMG. Due to the linear nature of state scale compared to the quadratic nature in transformers, the overall memory footprint and computational load of Mamba are significantly less for higher input sizes. Our proposed methodology introduces the use of the advantage of both the attention mechanism and SSM, by replacing that transformer block with a Vision Mamba (ViM)~\cite{zhu2024vision} unit for DMG.

\begin{figure*}[h!]
\centering
\includegraphics[width=18cm]{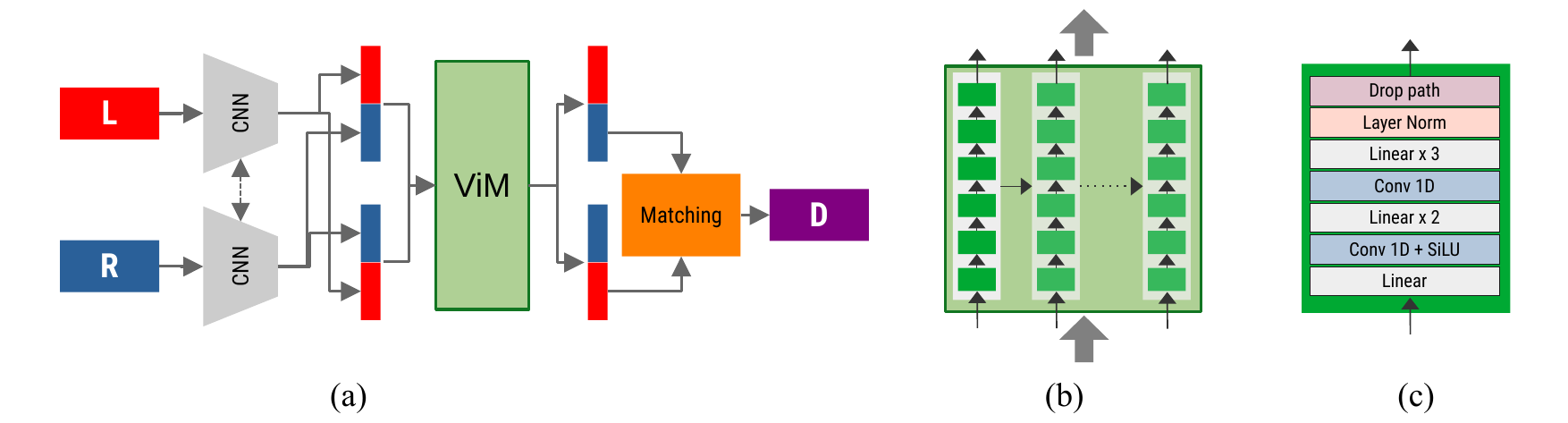}
\vspace{-3mm}
\caption{(a) Overview of the proposed architecture, (b) demonstrates the unrolling process across six Mamba encoders~\cite{gu2023mamba} (c) presents the internal structure of a Mamba encoders.}
\label{fig_proposed}
\end{figure*}



ViM is a model based on SSM Mamba~\cite{gu2023mamba} which can solve vision tasks. This selective state space memory reduces the parameter size of mamba from transformers, for an equivalent embedding size, restoring accuracy. ViM \cite{mambatransformervision} uses bidirectional mamba blocks to better incorporate global contextual information. The linear scalability of Mamba allows feasible resource budgets even while dealing with high-resolution images, unlike transformers, which scale quadratically. The basic idea of ViM starts from transforming a 2D image into a 1D encoded sequence to treat it like a linear state space. The transformation happens as $t \in \mathbb{R}^{H \times W \times C}$ into flattened 2D patches $x_p \in \mathbb{R}^{J \times (P^2 \cdot C)}$, where $(H, W)$ is the size of the input image, $C$ is the number of channels, and $P$ is the size of the image patches.\\

\vspace{-3mm}
\subsection{Proposed Architecture}
\vspace{-1mm}
The proposed architecture (Refer Figure~\ref{fig_proposed}) begins with a Convolutional Neural Network (CNN) encoder that extracts features from both the left and right stereo images at multiple scales. Let the features extracted from the left image be \( F_{\text{left}} \) and those from the right image be \( F_{\text{right}} \).These features are spatially structured, and when flattened, they are represented as tokens. Each token corresponds to a patch or region from the original image, capturing the local information of that region.

Our model incorporates a six-layer Mamba encoder, to handle multi-scale features. The Mamba encoder processes these features efficiently across scales. To retain spatial information, sine and cosine positional encodings \( P_{\text{left}} \) and \( P_{\text{right}} \) are added to the features of the left and right images, respectively:
\vspace{-1mm}
\begin{equation}
    F_{\text{left}} = F_{\text{left}} + P_{\text{left}}, \quad F_{\text{right}} = F_{\text{right}} + P_{\text{right}}
\end{equation}
\vspace{-1mm}
These positional encodings ensure that spatial relationships are preserved as the features are passed through the sequence-based ViM block. Before processing through the ViM block, the features \( F_{\text{left}} \) and \( F_{\text{right}} \) are flattened and concatenated symmetrically to facilitate joint learning between the stereo images. The left-to-right and right-to-left features are concatenated as follows:
\vspace{-1mm}
\begin{equation}
    \tilde{F}=\left[ F_{\text{left}}, F_{\text{right}} \right], \quad \tilde{F}_{\text{rev}}=\left[ F_{\text{right}}, F_{\text{left}}\right]
\end{equation}
\vspace{-1mm}
This symmetrical concatenation encourages cross-attention between the left and right image features, promoting implicit feature alignment. The concatenated representations \( \tilde{F} \) and \( \tilde{F}_{\text{rev}} \) are then passed into the ViM block for processing.

The concatenated features are processed layer by layer in the ViM block. For each layer \( l \), the feature update can be described as:
\vspace{-1mm}
\begin{equation}
    \tilde{F}_l=\mathcal{V}_{\text{block}}(\tilde{F}_{l-1}) + \tilde{F}_{l-1},  \quad\tilde{F}_{\text{rev}, l}=\mathcal{V}_{\text{block}}(\tilde{F}_{\text{rev}, l-1}) + \tilde{F}_{\text{rev}, l-1}
\end{equation}
where \( \mathcal{V}_{\text{block}} \) represents the transformation applied by the ViM block at the \( l \)-th layer. After processing through the ViM block, the output sequences \( \tilde{F}_{\text{output}} \) and \( \tilde{F}_{\text{rev, output}} \) are split to separate the left and right feature sequences. Since the concatenated features are symmetrical, we split them evenly:
\vspace{-1mm}
\begin{equation}
    F_{\text{left, output}} = \tilde{F}_{\text{output}}[:N], \quad F_{\text{right, output}} = \tilde{F}_{\text{output}}[N:]
\end{equation}
\vspace{-1mm}
where \( N \) is the length of the feature sequence for each image. Global and local feature matching is then performed on the separated left and right feature sequences \( F_{\text{left, output}} \) and \( F_{\text{right, output}} \) to estimate the disparity map. The disparity map is further refined using regressive refinement and upsampling, which applies convex sampling to smooth the final output. Finally, we retain only the positive disparities and upsample the predicted disparity map to the original image resolution.
\subsection{Matching}
 For depth estimation, with known camera parameters (K1, E1, K2, E2) for images I1 and I2, we discretize a predefined depth range into N candidates.For each depth candidate $d_{i}$, we compute 2D correspondences in $Frame2_{2}(F_{2})$ using the transformation H($\hat{G}_{2D}$) = K2E2E1\textsuperscript{-1}diK1\textsuperscript{-1}H($G_{2D}$), then perform bilinear sampling to obtain feature correlations $C = F1 · Fi2/\sqrt{D}$.These correlations are concatenated into $C_{d}$, normalized via softmax to get the matching distribution $M_{d}$, and the final depth is estimated as $V_{d}$ by computing the weighted average of $M_{d}$ with the depth candidates $G_{d}$.

\subsection{Proposed Performance Measure}
We also introduce a performance measure SOMER to jointly quantify the performance of DMG models in terms of accuracy, inference time and memory requirement. A DMG method needs to run fast while keeping the error rate low with low memory footprint, for the feasibility of deployment on edge devices. Thus of a DGM is proportional to FPS, and inversely proportional to error rate and the log of memory footprint. Hence, the higher the SOMER, the better is the model. As memory is of large range, for mapping and analyzing it to a compact form logarithm scale is employed. The measure is described as follows. 
\begin{equation}
\centering
    SOMER = \frac{FPS}{EPE \times log(M)}
\end{equation}
where the $M$ is the memory footprint.

\section{Datasets and Performance Measures}
We have run our analysis on some of the most popular datasets for stereo disparity benchmarking. They are briefly described as follows:\\
\noindent \textbf{KITTI}~\cite{geiger2013vision} is comprised of multiple sequences of urban video taken in different areas of Karlsruhe, Germany and the 2015 version that we used contains dynamic scenes with moving traffic containing $200$ scenes for training and testing with $20$ pair of frames per scene. The frames are $1242 \times 375$ in resolution.\\
\noindent \textbf{Sintel}~\cite{butler2012naturalistic} is a dataset from Max Plank Institute that comprises image pairs which are scenes from the movie (Sintel), hence having huge variations in environment type, weather, illumination, materials, structures, and scene dynamics. The scene frames are of dimension $1024 \times 436$ and consist of a total of $1628$ frames with $1064$ training image pairs and $564$ testing image pairs.\\ 
\noindent \textbf{Sceneflow}~\cite{mayer2016large} is a comparatively large dataset consisting of three sub-datasets: Flying Things, the dataset of random flying objects animated in 3D ($21818$ training frames and $4248$ test frames), Monkaa ($8591$ training frames only) a short movie like Sintel and Driving ($4392$ training frames only), a simulated driving dataset. The frame resolution is $1960 \times 540$. \\
\noindent \textbf{Virtual KITTI 2}~\cite{cabon2020virtual} is a dataset built by NeverLabs, inspired by the original KITTI dataset. It has a handsome collection of driving sequences. The resolution of the frames is $1242 \times 375$ and there are $446$ frames in each atmosphere and illumination variant. There are $6$ scene variants for a sequence and a total of $5$ sequences.\\
\noindent \textbf{Performance measures}: The metric of computation is End Point Error (EPE) and $3$ pixel error accuracy percentage for disparity (D1). EPE is the distance between estimated flow vectors and ground truth vectors. D1 error calculates the disparity pixels whose absolute error is greater than a threshold. For both EPE and D1 lower is better. We also used FPS (frame rate per second), memory footprint and SOMER for extensive analysis. 


\section{Experiments and Analysis}

\noindent \textbf{Implementation details:} All experiments are performed on a workstation environment comprising of Nvidia A6000 with AMD Ryzen $32$ core processor and $256$GB of RAM. The GPU has $10752$ cores and $48$GB of dedicated Virtual RAM. The model was trained on VKITTI2  for 80k iterations and then on Sceneflow for 10k iterations. The learning rate used was $1\times 10^{-5}$ and the optimizer was AdamW. \\
\begin{table}[]
    \scriptsize
    \centering
    \caption{Comparative performance over different datasets.}
    \vspace{-2mm}
    \label{tab:full_ablation}
    \begin{tblr}{
      colspec = {clccccc},
      row{1} = {font=\bfseries},
      hlines,
    }
    Dataset & Metric & Unimatch & I-GEV & Anynet & RAFT & Proposed \\ 
    \SetCell[r=5]{c} KITTI & EPE & 1.21 & \textbf{0.28} & 10.94 & 1.08 & 1.38 \\ 
    & D1 & \textbf{0.05} & 0.37 & 1.00 & 4.95 & 0.07 \\ 
    & FPS & 10.02 & 1.57 & 31.45 & 3.30 & \textbf{51.53} \\ 
    & MEM & 921 & 119 & 270 & \textbf{102} & 345 \\ 
    & SOMER & 1.21 & 1.173 & 0.513 & 0.66 & \textbf{6.409} \\ 
    \SetCell[r=5]{c} Sceneflow & EPE & 0.72 & \textbf{0.47} & 54.94 & 1.80 & 4.4 \\ 
    & D1 & 2.08 & 2.47 & 1.00 & 13.30 & \textbf{0.18} \\ 
    & FPS & 9.05 & 1.57 & 31.45 & 2.88 & \textbf{47.41} \\ 
    & MEM & 1007 & 119 & 154 & \textbf{102} & 375 \\ 
    & SOMER & 1.82 & 0.698 & 0.114 & 0.345 & \textbf{1.83} \\ 
    \SetCell[r=5]{c} Sintel & EPE & 1.45 & \textbf{0.32} & 88.04 & 0.45 & 11.53 \\ 
    & D1 & \textbf{0.04} & 1.18 & 0.99 & 1.31 & 0.24 \\ 
    & FPS & 10.55 & 1.87 & 42.22 & 3.52 & \textbf{52.53} \\ 
    & MEM & 882 & 118 & 263 & \textbf{101} & 334 \\ 
    & SOMER & 1.07 & 0.1223 & 0.086 & \textbf{1.692} & 0.785 \\ 
    \SetCell[r=5]{c} VKITTI2 & EPE & 1.95 & \textbf{0.92} & 88.55 & 0.92 & 1.146 \\ 
    & D1 & 0.13 & 5.70 & 0.99 & 6.33 & \textbf{0.06} \\ 
    & FPS & 10.04 & 2.42 & 42.24 & 1.60 & \textbf{50.62} \\ 
    & MEM & 922 & 118 & 275 & \textbf{102} & 346 \\ 
    & SOMER & 0.75 & 0.551 & 0.085 & 0.376 & \textbf{7.644} \\
    \SetCell[r=5]{c} All  & EPE(Avg) & 1.33 & \textbf{0.49} & 60.61 & 1.06 & 1.47 \\ 
    \SetCell[r=4]{c} datasets& D1(Avg) & 0.57 & 2.43 & 0.99 & 6.47 & \textbf{0.05} \\ 
    & FPS & 9.78 & 1.89 & 35.21 & 2.92 & \textbf{50.48} \\ 
    & MEM & 992 & 121 & 247 & \textbf{102} & 372 \\ 
    & SOMER & 1.06 & 0.8 & 0.105 & 0.595 & \textbf{5.8} \\
\end{tblr}
\end{table}

\begin{table}[ht]
\scriptsize
\centering
\caption{The statistics of speed of all methodologies. }\vspace{-2mm}
\begin{tblr}{
  width = \linewidth,
  colspec = {Q[187]Q[110]Q[171]Q[104]Q[131]Q[113]Q[96]},
  cell{2}{1} = {r=3}{},
  cell{5}{1} = {r=3}{},
  cell{8}{1} = {r=3}{},
  cell{11}{1} = {r=3}{},
  hline{1-2,5,8,11,14} = {-}{},
  hline{3-4,6-7,9-10,12-13} = {2-7}{},
}
Datasets  &      & Unimatch & IGEV & Anynet & RAFT & Proposed \\
KITTI     & min  & 9.94     & 0.63 & 26.66  & 3.12 &   \textbf{50.05}   \\
          & avg  & 10.02     & 1.57 & 31.45  & 3.30 &   \textbf{51.53}   \\
          & max  & 10.1     & 3.21 & 34.14  & 3.40 &     \textbf{51.93} \\
Sceneflow & min  & 8.9     & 0.68 & 26.66  & 2.54 &    \textbf{45.23}  \\
          & avg  & 9.05     & 1.57 & 31.45  & 2.88 &   \textbf{47.41}   \\
          & max  & 9.13     & 3.21 & 34.14    & 3.00 &    \textbf{48.63}  \\
Sintel    & min  & 10.46     & 0.82 & \textcolor{black}{39.70}    & 3.51 &   \textbf{50.49}   \\
          & avg  & 10.55     & 1.87 & \textcolor{black}{42.22}      & 3.52 &   \textbf{52.53}  \\
          & max  & 10.67    & 3.19 &   \textcolor{black}{42.8}   & 3.53 &    \textbf{53.58}  \\
v KITTI 2 & min  & 9.87     & 2.03 & \textcolor{black}{40.96}     & 1.25 &   \textbf{48.11}   \\
          & avg  & 10.04    & 2.42 & \textcolor{black}{42.24}      & 1.60 &  \textbf{50.62}    \\
          & max  & 10.14    & 2.97 & \textcolor{black}{42.76}    & 2.01 &      \textbf{51.45}
\end{tblr}
\vspace{1em}

\label{tab_fps}
\end{table}

\begin{table}[ht]
\scriptsize
\centering
\caption{Ablation study on different component of proposed model.}\vspace{-2mm}
\begin{tblr}{
    colspec = {Q[192]Q[179]Q[108]Q[137]Q[117]Q[181]},
    cell{2}{1} = {r=3}{},
    cell{5}{1} = {r=3}{},
    cell{8}{1} = {r=3}{},
    cell{11}{1} = {r=3}{},
    cell{14}{1} = {r=3}{},
    hline{1-2,5,8,11,14,17}={-}{},
    hline{3-4,6-7,9-10,12-13,15-16}={2-7}{},
}
Datasets & Metrics & Model w 1-pass w SA & Model w 2-pass w/o SA & Proposed model \\
KITTI & EPE & 2.69 & 4.09 & \textbf{1.38} \\
& D1 & 0.22 & 0.39 & \textbf{0.07} \\
& FPS & 51.09 & 33.55 & \textbf{51.7} \\
SceneFlow & EPE & 4.61 & 5.4 & \textbf{4.4} \\
& D1 & 0.21 & 0.28 & \textbf{0.18} \\
& FPS & 47.6 & 31.27 & \textbf{47.63} \\
Sintel & EPE & 11.71 & 12.12 & \textbf{11.53} \\
& D1 & 0.263 & 0.36 & \textbf{0.24} \\
& FPS & 52.39 & 34.34 & \textbf{52.6} \\
VKITTI2 & EPE & 5.2 & 6.46 & \textbf{1.146} \\
& D1 & 0.30 & 0.46 & \textbf{0.06} \\
& FPS & 50.13 & 33.17 & \textbf{51.23} \\
\end{tblr}
\label{tab:proposed_ablation}
\end{table}

\vspace{-5mm}
\noindent \textbf{Results:} The comparisons results of different methodologies are reported in Table~\ref{tab:full_ablation}. It must be noted that the proposed methodology worked well for most scenarios or was comparable to the best results for all datasets considering EPE, D1, Memory (MEM) and FPS. 

From the detailed analysis of FPS we can conclude that Unimatch~\cite{xu2023unifying}, RAFT-Stereo~\cite{teed2020raft}, I-GEV~\cite{xu2023iterative}, Any-net~\cite{chen2022improvement} are considerably less performant as compared to the proposed model. This performance disparity can be attributed to several factors, including the streamlined feature extraction process in our architecture, resulting in faster inference times. By leveraging advancements in the architecture via ViM, we achieve not only superior speed but also maintain competitive precision in terms of EPE and D1 across benchmark datasets, further solidifying our model's advantage in accurate real-time applications. Consequently, the proposed model sets a new standard in balancing accuracy and efficiency, making it a robust choice for resource-constrained environments where high FPS is critical. It can also be concluded from Table~\ref{tab_fps} that the proposed model performed superior for all scenarios of average(avg), minimum (min) and maximum (max) FPS compared to all datasets in comparison to the state-of-the-art. In this context, it is important to note that the FPS is computed by considering the time difference, before and after the call of the disparity computation function to negate any I/O overhead.

While our proposed model excels in terms of speed, its performance on the End-Point Error (EPE) and D1 metrics remains competitive with all state-of-the-art models, as can be seen in Table \ref{tab:full_ablation}. In particular, models such as Unimatch and I-GEV have slightly outperformed our approach in certain scenarios, demonstrating lower EPE and D1 values. This can be attributed to their more specialized refinement strategies and dense correlation computation, which offer higher precision in disparity estimation.

However, despite not achieving the best results in these metrics, our model remains highly competitive. It consistently produces results within a close range of the top performers, while maintaining a significantly lower computational overhead. This balance between accuracy and efficiency ensures that the proposed model is still highly suitable for real-time and large-scale applications, where speed is crucial, and a marginal trade-off in accuracy is acceptable. In environments where the need for fast and reliable processing outweighs the absolute minimization of EPE and D1, the proposed model presents a strong alternative to more computationally intensive approaches.

\begin{figure}[h!]
\centering
\includegraphics[width=1\columnwidth]{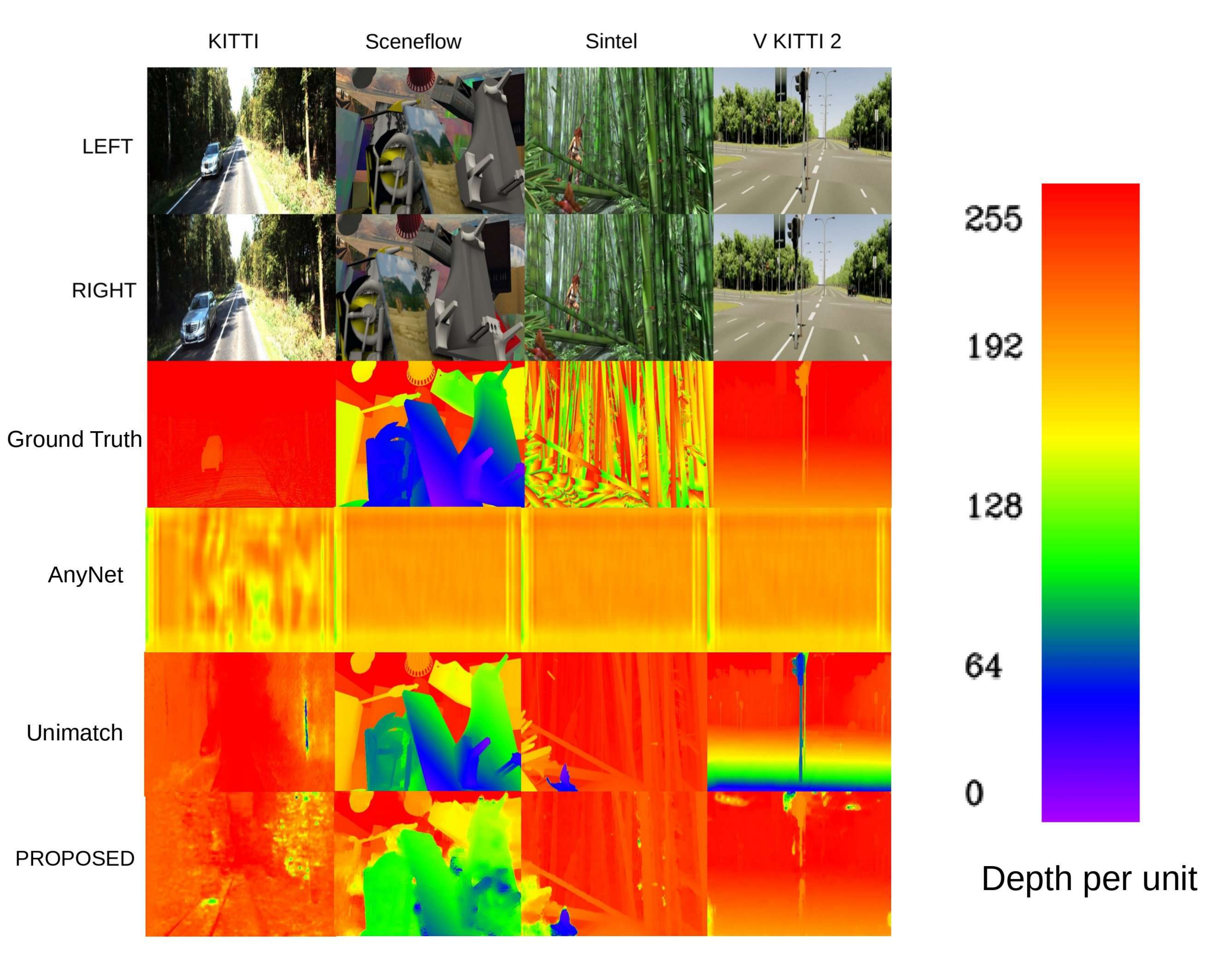}
\vspace{-8mm}
\caption{Different datasets used and their corresponding disparity heat maps. The heat maps are plotted in a rainbow spectrum with red being the lowest (farther away) and blue being the highest (nearer) disparity value. }
\label{hit}
\end{figure}

The memory footprint of different methodologies is reported in Table~\ref{tab:full_ablation}. Though memory requirement mainly depends on the parameter size of the model and should be agnostic to any type of data it can be observed that it varies when different datasets are considered. It can be observed that the memory footprint for the proposed model is much lower than Unimatch but higher than IGEV. Anynet and RAFT. To further analyse the proposed model with the state-of-the-art models and reach a concrete decision about its effectiveness for DMG in real-time scenarios, we analyse it with respect to the proposed SOMER matrix.

It can be noted from the Table~\ref{tab:full_ablation} that apart from Sintel, the proposed methodology performed better than the state-of-the-art models for SOMER. This concludes that the proposed model was able to dissolve the existing trade-off for a real-time and accurate model with low computation overhead for disparity map generation (DMG).   

The visualization of the disparity map from different datasets for the proposed model and the state-of-the-art models, along with left and right images, and ground-truths disparity are in Fig~\ref{hit}. It can be observed that the performance of the proposed model is quite near to Unimatch and ground truth, and better than AnyNet.

To evaluate the effects of various components of the proposed model we conducted an ablation study, the results of which are reported in Table \ref{tab:proposed_ablation}. The study involved changing the number of passes to ViM and adding self-attention to study their effects on the performance of our model in terms of EPE, D1 and FPS. Notably, adding an extra pass significantly decreases FPS, although adding self-attention has a negligible effect on the FPS and error rates.


\section{Conclusion}
\vspace{-2mm}
Disparity map generation (DMG) is a well-explored area in the robotic vision community. The literature on DMG exhibits that high accuracy and real-time generation with low computation overhead are found to be a trade-off. Further, most existing techniques aim to enhance the accuracy or speed of DMG. In contrast, this work attempted to extensively analyse the aforementioned trade-off of real timeliness vs accuracy of recent DMG methodologies on standard datasets such as KITTI, Virtual KITTI, Sceneflow and Sintel. Concluding from this we propose a ViM-based model which can bridge the gap of time, memory requirement and accuracy of a real-time DMG. Moreover, we introduce a new measure for the overall joint analysis of the efficacy of DGM models by considering accuracy, speed and memory footprint. The results and analysis conclude that we were able to dissolve the gap in speed, accuracy and memory better than any other state-of-the-art techniques \textcolor{black}{as per our proposed measure for DMG}. 

\section*{Acknowledgment}
This work was partially funded by BITS Pilani New Faculty Seed Grant (NFSG/HYD/2023/H0831).

\bibliographystyle{IEEEtran}
\bibliography{IEEEabrv,egbib}

\end{document}